\documentclass{article}

\usepackage[preprint]{neurips_2025}

\usepackage[pagebackref,breaklinks,colorlinks]{hyperref}
\usepackage[utf8]{inputenc} 
\usepackage[T1]{fontenc}    
\usepackage{hyperref}       
\usepackage{url}            
\usepackage{booktabs}       
\usepackage{amsfonts}       
\usepackage{nicefrac}       
\usepackage{microtype}      
\usepackage{enumitem}

\usepackage{graphicx}
\usepackage{booktabs}

\usepackage[accsupp]{axessibility}  

\usepackage{graphicx}

\usepackage{tikz}
\usepackage{comment}
\usepackage{amsmath,amssymb} 
\usepackage{bbm}
\usepackage{booktabs}
\usepackage{graphicx}
\usepackage{subfig}
\usepackage{bm}
\usepackage{multirow}
\usepackage{ulem} 

\usepackage{algorithm}

\usepackage{mdframed}
\usepackage{mdframed}
\usepackage{amsmath, amssymb}

\usepackage{orcidlink}

\usepackage{amstext}
\usepackage{amssymb}
\usepackage{booktabs}

\usepackage{tikz}
\usepackage{comment}
\usepackage{amsmath,amssymb} 
\usepackage{bbm}
\usepackage{booktabs}
\usepackage{graphicx}
\usepackage{bm}
\usepackage{multirow}
\usepackage{cases}
\usepackage{color}
\usepackage{ulem}
\definecolor{dino}{RGB}{249,231,227}

\usepackage{url}
\usepackage{epsfig}
\usepackage{amsmath}
\usepackage{amssymb}
\usepackage{amsthm}
\usepackage{multirow}
\usepackage{soul}
\usepackage{footnote}
\usepackage{tablefootnote}
\usepackage{caption}
\usepackage{subcaption}
\usepackage{mathtools}

\usepackage{url}
\usepackage{multirow}
\usepackage{booktabs}
\usepackage{pifont}
\usepackage{graphicx}
\usepackage{amsmath}
\usepackage{algorithm}
\usepackage{mathtools}
\usepackage{algorithm}
\usepackage{algpseudocode}
\usepackage{listings}
\usepackage{wrapfig}
\usepackage{subcaption}

\definecolor{top1}{RGB}{255,179,179}
\definecolor{top2}{RGB}{255,217,179}
\definecolor{top3}{RGB}{255,255,179}

\usepackage{tikz}
\usepackage{comment}
\usepackage{amsmath,amssymb} 
\usepackage{bbm}
\usepackage{booktabs}
\usepackage{graphicx}
\usepackage{bm}
\usepackage{multirow}

\usepackage{xcolor,pifont}
\newcommand*\colourcheck[1]{%
  \expandafter\newcommand\csname #1check\endcsname{\textcolor{#1}{\ding{52}}}%
}
\colourcheck{blue}
\colourcheck{green}
\colourcheck{red}
\usepackage{makecell}

\title{
EmbodieDreamer: Advancing Real2Sim2Real Transfer for Policy Training via Embodied \\ World Modeling
}

\author{
    Boyuan Wang\textsuperscript{\rm 1, 2}\footnotemark[1]~,
    Xinpan Meng\textsuperscript{\rm 1, 2}\footnotemark[1]~,
    Xiaofeng Wang\textsuperscript{\rm 1, 2}\footnotemark[1]~, 
    Zheng Zhu\textsuperscript{\rm 1}\footnotemark[1]~\footnotemark[2]~~,\\
    \textbf{Angen Ye}\textsuperscript{\rm 1, 2},
    \textbf{Yang Wang}\textsuperscript{\rm 1},
    \textbf{Zhiqin Yang}\textsuperscript{\rm 1},
    \textbf{Chaojun Ni}\textsuperscript{\rm 1,3},
    \textbf{Guan Huang}\textsuperscript{\rm 1}, 
    \textbf{Xingang Wang}\textsuperscript{\rm 2}\footnotemark[2]\\
    \textsuperscript{\rm 1}GigaAI~~
    \textsuperscript{\rm 2}Institute of Automation, Chinese Academy of Sciences~~
    \textsuperscript{\rm 3}Peking University~~
    \\
    \url{https://embodiedreamer.github.io/}
}
\begin{document}

\maketitle
\renewcommand{\thefootnote}{\fnsymbol{footnote}}
\footnotetext[1]{
These authors contributed equally to this work. 
}
\footnotetext[2]{\mbox{Corresponding authors. zhengzhu@ieee.org, xingang.wang@ia.ac.cn.}}

\begin{abstract}
        
The rapid advancement of Embodied AI has led to an increasing demand for large-scale, high-quality real-world data. However, collecting such embodied data remains costly and inefficient. As a result, simulation environments have become a crucial surrogate for training robot policies. Yet, the significant Real2Sim2Real gap remains a critical bottleneck, particularly in terms of physical dynamics and visual appearance. To address this challenge, we propose \textit{EmbodieDreamer}, a novel framework that reduces the Real2Sim2Real gap from both the physics and appearance perspectives. 
Specifically, we propose \textit{PhysAligner}, a differentiable physics module designed to reduce the Real2Sim physical gap. It jointly optimizes robot-specific parameters such as control gains and friction coefficients to better align simulated dynamics with real-world observations. In addition, we introduce \textit{VisAligner}, which incorporates a conditional video diffusion model to bridge the Sim2Real appearance gap by translating low-fidelity simulated renderings into photorealistic videos conditioned on simulation states, enabling high-fidelity visual transfer.
Extensive experiments validate the effectiveness of \textit{EmbodieDreamer}. The proposed \textit{PhysAligner} reduces physical parameter estimation error by 3.74\% compared to simulated annealing methods while improving optimization speed by 89.91\%. Moreover, training robot policies in the generated photorealistic environment leads to a 29.17\% improvement in the average task success rate across real-world tasks after reinforcement learning. Code, model and data will be publicly available.

\begin{figure}[ht]
    \centering
    \includegraphics[width=\textwidth]{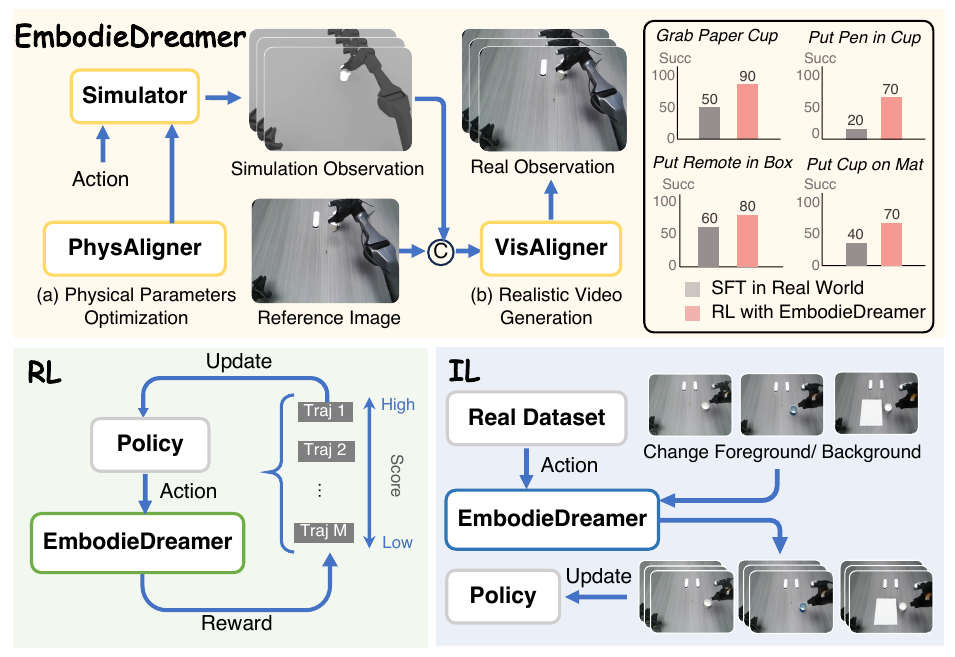}
    \caption{
        \textit{EmbodieDreamer} is a unified Real2Sim2Real framework that incorporates \textit{PhysAligner} for rapid physical parameter optimization from real observations and \textit{VisAligner} for generating visually realistic scenes. 
        Furthermore, \textit{EmbodieDreamer} supports RL training of policy models through preference learning based on trajectory evaluation, and facilitates IL training by generating diverse, unseen observations.
        Models trained within \textit{EmbodieDreamer} show significantly better performance compared to policies fine-tuned on real-world data.
    }
    \label{fig: main_demo}
\end{figure}

\end{abstract}
\section{Introduction}

Recent advances in Embodied AI have driven remarkable progress in robot learning, enabling agents to perform complex manipulation, navigation, and decision-making tasks in real-world environments. Central to this progress is the availability of large-scale, high-quality embodied data that captures both the sensory input and motor interaction of agents operating in diverse physical settings. However, collecting such data in the real world is expensive, time-consuming, and often limited in scalability and reproducibility. These limitations motivate the growing reliance on simulation environments \cite{mujuco,rlbench,sapien,simplerenv,genesis,vlabench,roboverse,robotwin} as a practical alternative for training and evaluating robotic policies.

Despite their practical utility, existing simulation environments suffer from the Real2Sim2Real discrepancy. Two key sources of this discrepancy are differences in physical dynamics and visual appearance. 
First, the physical behaviors simulated in virtual environments often diverge from real-world dynamics due to oversimplified assumptions or unmodeled parameters such as joint stiffness, damping, or arm friction. These mismatches result in inaccurate motion trajectories and undermine the reliability of simulation-based control.
Second, the rendered appearance of simulated scenes typically lacks the visual fidelity required to generalize to real-world sensory inputs. As a result, perception models trained in simulation may fail to operate effectively in real environments.
Together, these inconsistencies pose significant challenges for deploying policies trained in simulation, especially for tasks demanding precise physical interaction or high-quality visual perception.

To better align simulation with the real world in both dynamics and appearance, we introduce \textit{EmbodieDreamer}, a unified framework for improving the fidelity of simulation environments through embodied world modeling. Our approach comprises two complementary components designed to address the Real2Sim2Real discrepancy from both the physical and visual perspectives.
First, to mitigate the Real2Sim gap in physical dynamics, we propose \textit{PhysAligner}, a differentiable physics optimization module that estimates robot-specific parameters such as control gains and friction coefficients. In contrast to the simulated annealing-based optimization used in systems like SimplerEnv \cite{simplerenv}, \textit{PhysAligner} leverages gradient-based optimization enabled by differentiable physics to more efficiently and accurately align simulated behaviors with real-world trajectories.
Second, to address the Sim2Real discrepancy in visual appearance, we introduce \textit{VisAligner}, a conditional video diffusion model~\cite{svd,cogvideo,humandreamer,humandreamerx} that synthesizes photorealistic RGB videos from simulation inputs. Unlike prior systems such as Rebot \cite{rebot}, RoboEngine \cite{roboengine}, IRASim \cite{irasim}, which primarily modify either the robot or the foreground/background appearance in isolation, \textit{VisAligner} explicitly disentangles the foreground, background, and robot components. This structured conditioning enables coherent and realistic video generation, making it suitable as a high-fidelity simulation environment for downstream policy learning tasks, including imitation learning \cite{huang2025dexterous,zare2024survey}and reinforcement learning \cite{tang2025deep,lin2025sim}.

We conduct extensive experiments to evaluate the effectiveness of \textit{EmbodieDreamer}. Our results show that \textit{PhysAligner} significantly improves the alignment between simulated and real-world dynamics, reducing physical parameter estimation error by 3.74\% and accelerating optimization by 89.91\% compared to simulated annealing baselines. Furthermore, we demonstrate that training policies within the generated photorealistic environment results in a 29.17\% improvement in the average real-world task success rate after reinforcement learning, confirming the practical benefits of our approach for robot policy learning.

Our contribution can be summarized as follows:
\begin{itemize}
    \item We propose \textit{EmbodieDreamer}, a unified framework that advances Real2Sim2Real transfer for policy training by leveraging embodied world modeling to enhance both physical and visual fidelity in simulation environments.
    
    \item We develop \textit{PhysAligner}, a differentiable physics optimization module designed to reduce the Real2Sim physical gap by accurately estimating robot-specific parameters such as control gains and friction coefficients, thereby aligning simulated dynamics with real-world behavior.
    
    \item We introduce \textit{VisAligner}, a conditional video diffusion model that bridges the Sim2Real appearance gap by translating low-fidelity simulation renderings into photorealistic videos through disentangled modeling of robot, foreground, and background elements.
    
    \item We conduct extensive experiments across multiple benchmarks to validate the effectiveness of \textit{EmbodieDreamer}, demonstrating improved parameter estimation accuracy, faster convergence compared to baseline methods, and higher task success rates in both reinforcement learning and imitation learning policy training.
\end{itemize}

\section{Related Work}

\subsection{Robot Policy Models}

In recent years, significant breakthroughs have propelled the field of robotic manipulation policies forward, driven by innovations in Vision-Language-Action (VLA) models \cite{pi0,pi05,team2024octo}, diffusion-based frameworks \cite{carvalho2023motion,ze20243d}, and preference optimization techniques \cite{rafailov2023direct, grape}. Chronologically, ACT \cite{zhao2023learning} introduced transformer-based action chunking strategies for robotic manipulation. RT-2 \cite{brohan2023rt} pioneered the integration of large language models into VLA architectures through semantic grounding. Strategies based on diffusion models, including Diffusion Policy \cite{chi2023diffusion}, RDT-1B \cite{rdt1b}, DP3 \cite{ze20243d}, and RDP \cite{xue2025reactive}, have showcased their ability to generate multimodal actions, facilitating generalization across tasks, embodiments, and modalities. Robotic policies leveraging large-scale, publicly available dataset pre-trained models, such as OpenVLA \cite{openvla}, Octo \cite{octo}, and $\pi_0$ \cite{pi0}, demonstrate generalizability by fine-tuning on specific tasks, exhibiting impressive transferability across various robotic platforms. GRAPE \cite{grape} enhances various VLA architectures through trajectory-level preference optimization, utilizing success/failure trials to improve out-of-distribution generalization while supporting customizable objectives like safety and efficiency. Furthermore, recent developments include layered architectures like DexGraspVLA \cite{zhong2025dexgraspvla} and the incorporation of chain-of-thought reasoning in Cot-vla \cite{zhao2025cot}, which enhance VLA models’ ability to handle complex, long-horizon tasks more effectively. These advances underscore the growing dependence of policy models on high-quality, diverse data and realistic simulation environment to achieve robust and generalizable performance in real-world settings.

\subsection{Simulation Environment}
Effective simulation environments underpin the scalable development and evaluation of both policies and world models. MuJoCo \cite{mujuco} provides highly accurate physics simulations and remains a cornerstone for reinforcement learning. Meanwhile, platforms like RLBench \cite{rlbench} and VLABench \cite{vlabench} emphasize task diversity and parallel data collection, enabling efficient exploration of complex manipulation scenarios. SAPIEN \cite{sapien} and Genesis \cite{genesis} leverage GPU acceleration to achieve high-performance rendering and scalable reinforcement learning, significantly improving training throughput for large-scale environments. SimplerEnv \cite{simplerenv} , alongside these frameworks, offers a lightweight and extensible collection of manipulation tasks mapped closely to real robot setups, such as the WidowX BridgeV2 \cite{bridgedatav2} and Google Robot \cite{rt1}, facilitating rapid iteration and reproducible comparative studies. Furthermore, RoboVerse \cite{roboverse} proposes a unified platform supporting seamless transitions between multiple simulators, allowing hybrid integration to harness complementary strengths across environments. RoboTwin \cite{robotwin} presents a digital-twin framework that aligns twin robot and virtual agent behaviors via 3D generative modeling and large-language-model-driven data augmentation, thereby facilitating high-fidelity sim-to-real transfer on collaborative platforms. Together, these simulation platforms provide the necessary infrastructure for training, validating, and stress-testing advanced control policies and world models, accelerating progress toward robust, generalizable robotic systems.
However, current simulation environments still face critical challenges including inaccurate physical parameters, limited visual realism, and significant sim-to-real transfer gaps that degrade policy model performance in real-world deployments.

\subsection{Embodied World Models}
The development of world models has seen significant advancements in recent years, with notable progress reported in \cite{magicdrive, drivedreamer,worlddreamer,gaia, drivedreamer4d, streetgaussian, recondreamer}. Building on these developments, robot world models aim to provide an internal representation of the environment that supports action simulation and prediction, thereby enhancing the robot's decision-making and planning capabilities in real-world scenarios.
ACDC \cite{acdc} and UniPi \cite{unipi} unify affordance prediction and physics forecasting within a single multilingual, multimodal video model, enabling policy learning directly from RGB demonstrations across diverse manipulation benchmarks; subsequently, IRASim \cite{irasim} leverages generative models to produce highly realistic robot videos from a single initial frame, offering an interactive simulation proxy for real-robot learning. Building on this foundation , RoboDreamer \cite{robodreamer} employs self-supervised multi-stage diffusion to decompose natural language commands into primitive actions and synthesize latent videos for combinatorial generalization, while ManipDreamer \cite{manipdreamer} enhances this framework by incorporating hierarchical action trees and semantic visual adapters to strengthen instruction relationships and improve temporal and visual consistency. RoboTransfer \cite{robotransfer} introduces a diffusion-based video generation framework that enables geometry-consistent multi-view video synthesis with fine-grained control over scene components, bridging the sim-to-real gap through explicit 3D priors and cross-view conditioning. With advancements in 3D reconstruction methods~\cite{3dgs,nerf}, several studies~\cite{drivedreamer4d,recondreamer,humandreamerx,recondreamerplus,wonderturbo,zhou2025gs,10902412,wang2024og} have leveraged these techniques to bridge the reality gap in various fields, including robotics. Robo-GS \cite{robo-gs} presents a differentiable Real2Sim pipeline that integrates mesh geometry with 3D Gaussian splattings (3DGS) \cite{3dgs} for high-fidelity digital asset reconstruction, ReBot \cite{rebot} expands visual-language-action datasets via real-to-sim-to-real trajectory replay and background inpainting, and SplatSim \cite{splatsim} adopts 3DGS instead of traditional mesh rendering to generate photorealistic synthetic data that significantly narrows the sim-to-real gap.

\section{Method}
\subsection{Framework Overview}

\begin{figure}[ht]
    \centering
    \includegraphics[width=\textwidth]{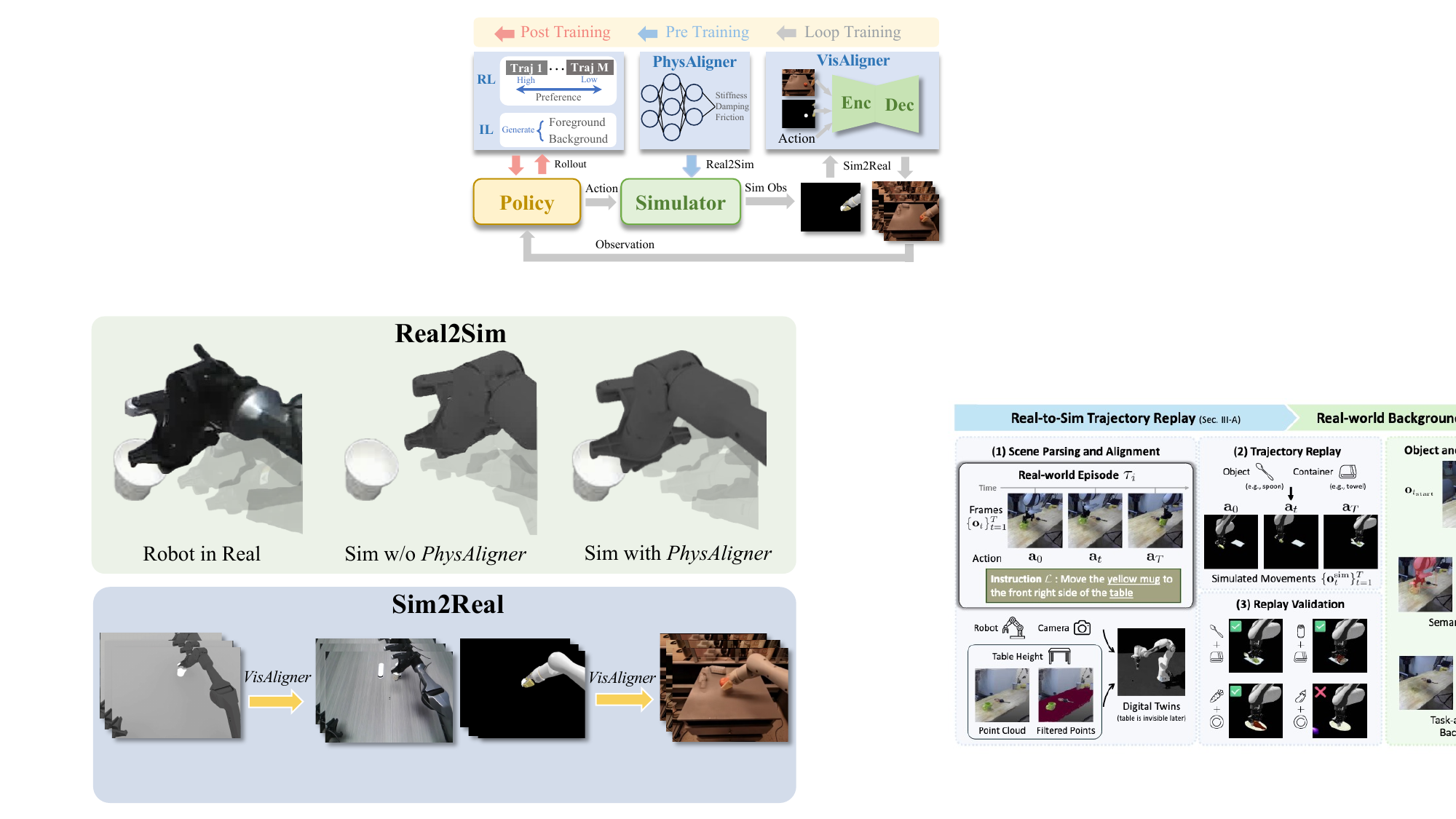}
    \caption{
    \textit{EmbodieDreamer} framework integrates \textit{PhysAligner} and \textit{VisAligner} to reduce the Real2Sim2Real gap in physics and appearance. \textit{PhysAligner} optimizes simulator dynamics, while \textit{VisAligner} translates simulated renderings into realistic observations for robot policy training.
    }
    \label{fig:framework}
\end{figure}

We present \textit{EmbodieDreamer}, a robot policy enhancement framework that integrates two core components: \textit{PhysAligner} and \textit{VisAligner}. This framework effectively reduces the Real2Sim2Real gap in both physical dynamics and visual appearance. As illustrated in Figure~\ref{fig:framework}, \textit{PhysAligner} consists of a differentiable network for optimizing physical parameters, enabling the simulator to better approximate real-world dynamics. \textit{VisAligner} leverages a video diffusion model to translate simulated renderings into more realistic visual observations, which are then fed into the robot policy model to generate actions that drive the simulator. Through this pipeline, \textit{EmbodieDreamer} functions as a realistic and adaptable simulation environment for robot policy training, supporting both imitation learning and reinforcement learning paradigms.

\subsection{PhysAligner}

\begin{figure}[ht]
    \centering
    \includegraphics[width=\textwidth]{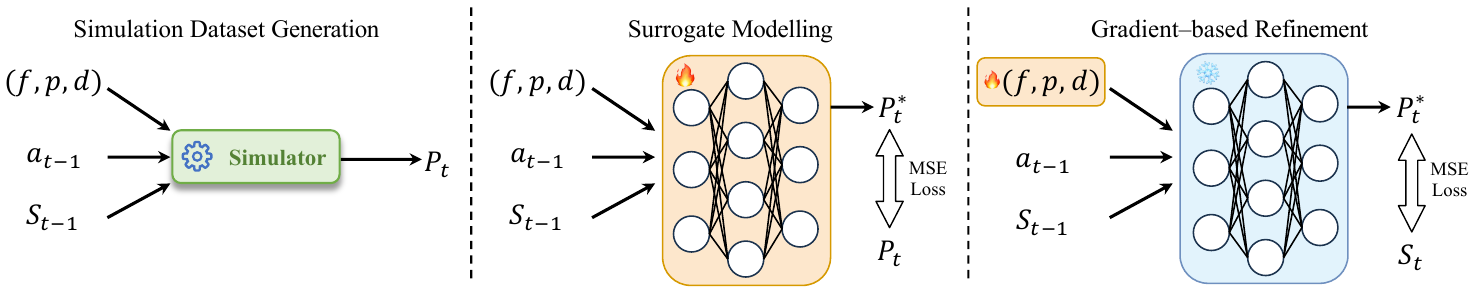}
    \caption{The figure illustrates the workflow of \textit{PhysAligner}. First, a large amount of data is generated using a simulator. Then, a surrogate model is trained to fit the data. Finally, the physical parameters are optimized through gradient descent.}
    \label{fig:sysid}
\end{figure}

Under Real2Sim settings, a physical gap inevitably exists between the simulated and real-world dynamics. When applying identical action signals to both the simulator and the physical robot, our objective is to minimize the control discrepancy between the two domains. Ideally, open-loop execution of motion sequences in the simulator should yield trajectories that closely match those observed in the real world. However, achieving this alignment requires accurate specification of system parameters, such as control gains, which are often determined empirically or through extensive trial and error.

SimplerEnv \cite{simplerenv} employs simulated annealing to optimize these parameters, while conceptually effective, this stochastic search method entails high computational cost and frequently suffers from slow or unstable convergence. To address these limitations, we propose a more efficient framework based on differentiable, learning-driven optimization. This approach enables gradient-based parameter estimation and greatly enhances both convergence speed and stability.

As shown in Figure \ref{fig:sysid}, the system identification pipeline consists of three sequential stages:

\textbf{Simulation Dataset Generation.}
We use $(f, p, d)$ to denote the friction, stiffness, and damping coefficients in the simulator. At time step $t-1$, the action sequence $a_{t-1}$ and corresponding state $S_{t-1}$, which includes camera observation and proprioception, are extracted from real-world collected data. We then randomly sample multiple $(f,p,d)$ pairs. The simulator executes the given action $a_{t-1}$ under the initial state condition $S_{t-1}$, and acquires the next state:  
\begin{equation}
    P_t = \text{Sim}_{(f,p,d)}(a_{t-1}, S_{t-1}).
\end{equation}

\textbf{Surrogate Modeling.}
Next, we construct a surrogate model $M_{f,p,d}$ to learn the relationship between physical parameters and system state transitions. The model takes as input the physical parameters, actions, and states from the generated simulation dataset, and produces the following prediction:
\begin{equation}
    P_t^* = M_{f,p,d}(a_{t-1},S_{t-1}),
\end{equation}

The model is trained by minimizing the mean-squared error between predicted and simulated trajectories, thereby constructing a differentiable approximation of the simulator’s non-smooth dynamics:

\begin{equation}
   \mathcal{L}_{\text{surrogate}} = \frac{1}{T} \sum_{t=1}^T \|P_t^{\ast} - \hat{P}_t\|^2.
\end{equation}

\textbf{Gradient-based Refinement.}
Finally, the surrogate model is used to refine the physical parameters via gradient descent. With the network weights fixed, the $(f,p,d)$ parameters are iteratively updated to minimize the following loss function:

\begin{equation}
    \mathcal{L}_{\text{para}} = \frac{1}{T} \sum_{t=1}^T \|P_t^{\ast} - S_t\|^2.
\end{equation}

This iterative process continues until the simulated output aligns closely with the observed trajectories, yielding an optimized set of PD controller parameters $(f^*,p^*,d^*)$ that best reproduce real-world dynamics. More details about loss function are provided in Appendix \ref{app:physaligner}.

\subsection{VisAligner}
\begin{figure}[ht]
    \centering
    \includegraphics[width=\textwidth]{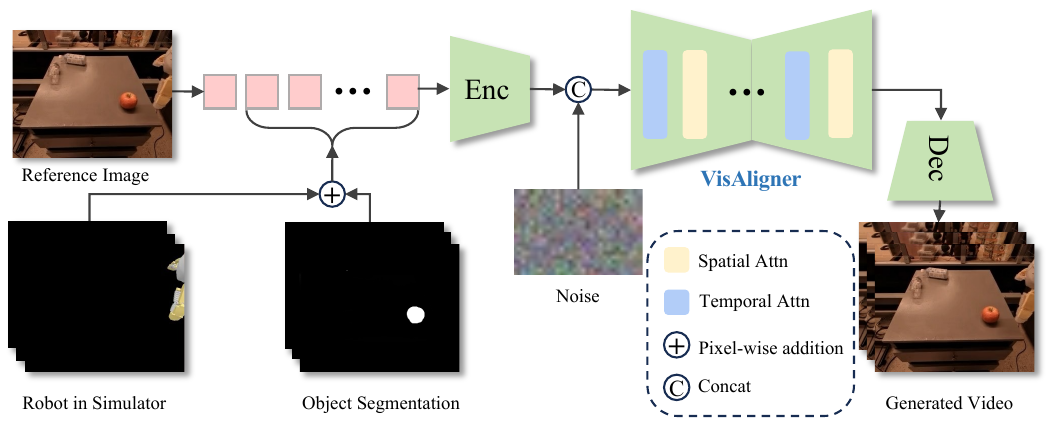}
    \caption{The framework of \textit{VisAligner}. A reference image containing the initial background and robot appearance information serves as the first frame of the conditioned video. The subsequent frames are generated by performing pixel-wise addition of the robot's motion observations from the simulated environment and the segmentation masks of the foreground objects. These frames are then encoded into latents via a VAE encoder~\cite{vae}, concatenated with noise along the channel dimension, and input to \textit{VisAligner} for denoising. Spatial-temporal attention mechanisms are employed to capture long-range dependencies across both spatial and temporal dimensions, thereby enhancing the coherence and visual quality of the generated video. The final video is obtained by decoding the denoised latents.}
    \label{fig:visaligner}
\end{figure}

To address the appearance gap in Sim2Real transfer, we employ a video generation framework and propose \textit{VisAligner} to achieve appearance transfer, as illustrated in Figure~\ref{fig:visaligner}. Given a manually collected episode \(e\) of length \(f\), we denote the joint positions and actions as $q, a \in \mathbb{R}^{f \times 14}$, 
and the ground-truth video captured by the camera as $V_{\mathrm{gt}} \in \mathbb{R}^{f \times H \times W \times c}.$

We construct a training batch \(\{\,(q^{(i)}, a^{(i)}, V_{\mathrm{gt}}^{(i)})\}_{i=1}^N\) to optimize \textit{VisAligner}.
For the conditioning inputs, we decompose control into three components: the robot, the foreground object, and the background. 

First, for the robot, we utilize its URDF model \(u_r\) and replay the joint trajectories \(q\) in the simulator:
\begin{equation}
    \mathrm{obs}_r = \operatorname{Sim}(q, u_r),
\end{equation}

where the virtual camera parameters are manually calibrated to align the simulated manipulator with its real-world counterpart. 

Second, for the foreground object, we apply GroundSAM \cite{grounding, sam2} to segment the object and obtain a binary mask $\mathrm{seg}$. This mask is added to the original observation image $\mathrm{obs}_r$ in a pixel-wise manner to generate the simulation condition signal $\mathrm{c}_{\mathrm{sim}}$.
Third, as the camera remains static, the background is invariant. Therefore, a single reference image $I_r \in \mathbb{R}^{H \times W \times c}$ is used as the background control signal. During training, the conditioning sequence for a video consisting of $f$ frames is formed by concatenating $I_r$ with the last $f-1$ frames of $\mathrm{c}_{\mathrm{sim}}$, denoted as:  
$\mathrm{Condition} = [I_r, \mathrm{c}_{\mathrm{sim}}^{t_1}, ..., \mathrm{c}_{\mathrm{sim}}^{t_{f-1}}]$.

We adopt a latent video diffusion model \cite{svd} as the backbone. Specifically, the conditional simulation frames $\mathrm{cond}_{\mathrm{sim}}$ and the ground‑truth video $V_{\mathrm{gt}}$ are each encoded via a Variational Autoencoder (VAE) \cite{vae} encoder $\varepsilon(\cdot)$, resulting in latent representations
$y_{\mathrm{c}}, z_{\mathrm{0}} \in \mathbb{R}^{f\times h\times w\times c}.$ 
At diffusion timestep $t$, the concatenated latent $z$ is corrupted as:
\begin{equation}
z_t = \sqrt{\alpha_t}z_0 + \sqrt{1-\alpha_t}\epsilon,
\quad \epsilon \sim \mathcal{N}(0,I),
\end{equation}

where $\{\alpha_t\}_{t=1}^T$ is a predefined noise schedule follow \cite{svd}. The denoising network $\epsilon_\theta$ is optimized by minimizing the following loss:

\begin{equation}
\mathcal{L}
=
\mathbb{E}_{z_0,y_\mathrm{c}} \| z_0 -\epsilon_\theta(z_t,t,y_\mathrm{c}) \| ^2.
\end{equation}
This objective encourages the network to predict $z_0$, enabling faithful reconstruction of the video. Empirically, conditioning on both the simulated mask and real video latents leads to significant improvements in visual consistency and appearance alignment across frames.

\section{Experiments}
We utilize the Cobot Mobile ALOHA, a robotic platform grounded in the Mobile ALOHA system design \cite{mobile_aloha}, produced by AgileX.AI, to support manual data collection and testing. For each task, 50 episodes are collected. We conduct a series of experiments including evaluation of system identification performance, quantitative assessment of generation quality, training of the policy model with both reinforcement learning and imitation learning, and finally, evaluation of task success rates on the real robot.

\subsection{Quantity result Comparison of \textit{PhysAligner}}

\paragraph{Experimental Setup.}
We conduct experiments on 20 randomly selected episodes from the RT-1 dataset \cite{rt1}. Our method trains a three-layer multilayer perceptron (MLP) as a surrogate model to approximate the system dynamics, taking physical parameters—including stiffness, damping, and friction—as input. In contrast to the baseline SimplerEnv \cite{simplerenv}, which employs 400 simulated annealing steps to optimize PD gains, we randomly sample 50 parameter sets for initial training and subsequently refine the MLP via gradient descent with a learning rate of 0.001. All other configurations, including data normalization, are kept consistent to ensure a fair comparison. Performance is evaluated using the mean squared error (MSE) between the predicted and ground-truth end-effector trajectories, and we additionally report the total computation time for data preparation and parameter optimization.

\paragraph{Experimental Results.}
As shown in Table~\ref{tab:comp_sysid}, \textit{PhysAligner} achieves a lower overall trajectory error, compared to the baseline SimplerEnv, mainly due to a significant reduction in rotation error. Although translation error slightly increases, the method demonstrates approximately 10× speedup in total computation time, reducing optimization from over 12,800 seconds to around 1,300 seconds.

These results highlight that the differentiable surrogate model effectively approximates system dynamics with minimal accuracy loss, achieving a better trade-off between estimation performance and computational efficiency. This makes the proposed method more suitable for practical system identification tasks.

\begin{table}[htbp]
\centering
\caption{Performance comparison with the SimplerEnv on system identification tasks.}
\label{tab:comp_sysid}
\resizebox{0.9\linewidth}{!}{
\begin{tabular}{lcccc}
  \toprule
  Method & Trajectory error &Rotation error&Translation error & Time cost(s) \\
  \midrule
  SimplerEnv & 0.2245 &0.1282&0.0963&12888.43\\
  \textit{PhysAligner}(Ours) & \textbf{0.2161} & \textbf{0.1101} & \textbf{0.1060} & \textbf{1299.92}\\
  \bottomrule
\end{tabular}%
}
\end{table}

\subsection{Video Generation Quality Comparison of VisAligner}

\paragraph{Experimental Setup.}
We conduct experiments on the RT-1 dataset \cite{rt1}, where we randomly split the data into a training set of 4,762 samples and a test set of 100 samples. Our model, \textit{VisAligner}, is built upon the SVD \cite{svd} with latent feature channels set to $ c = 4 $. The input resolution is fixed at $ 640 \times 480 $, with a video length of 25 frames.

To evaluate the impact of object mask conditioning on video generation quality, we perform ablation studies by comparing models trained with and without object segmentation inputs. We use standard image and video similarity metrics——FVD \cite{fvd} , PSNR, SSIM and LPIPS \cite{lpips} ——to quantitatively assess the fidelity and temporal coherence of the generated videos.

\paragraph{Experimental Results.} The results, summarized in Table~\ref{tab:comp_gen}, clearly demonstrate the effectiveness of our approach in generating high-quality videos when incorporating object mask information as a condition.

\begin{table}[htbp]
 \centering
\caption{Generation quality comparison.}
\label{tab:comp_gen}
\resizebox{0.7\linewidth}{!}{
\begin{tabular}{lcccc}
  \toprule
  Method & FVD$\downarrow$ & PSNR$\uparrow$ & SSIM$\uparrow$ & LPIPS$\downarrow$\\
  \midrule
  w/o object segmentation & 422.73 & 23.067 & 0.7939 & 0.0713\\
  \textit{EmbodieDreamer} (ours) &   \textbf{176.27} & \textbf{24.518} & \textbf{0.821} & \textbf{0.0554} \\
  \bottomrule
\end{tabular}
}
\end{table}

\begin{figure}[ht]
    \centering
    \includegraphics[width=\textwidth]{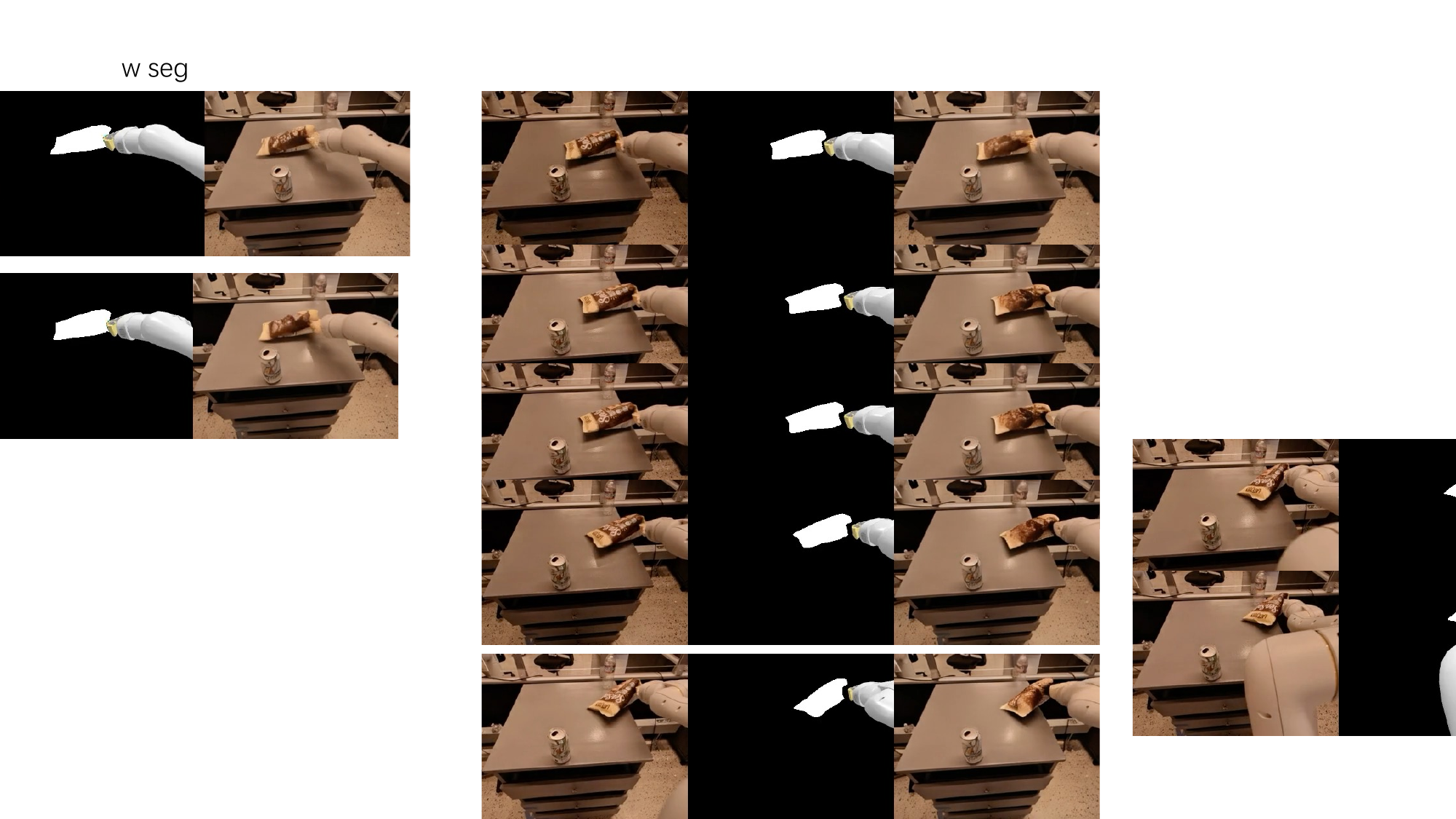}
    \caption{The visualization comparison of whether foreground object segmentation is used as a conditioning input. The four rows show: (1) Ground Truth (GT), (2) generated without segmentation condition, (3) segmentation map used as condition, and (4) generated result with segmentation condition.}
    \label{fig:comp_gen}
\end{figure}

The visualization comparison in Figure~\ref{fig:comp_gen} demonstrates that models without object segmentation exhibit deformation or disappearance of target objects during grasping, while segmentation-conditioned outputs maintain structural integrity and pose consistency.

\subsection{RL with EmbodieDreamer}

\paragraph{Experimental Setup.}

We adopt the supervised fine-tuning (SFT) of the ACT model \cite{act} as our baseline, where each task is trained with 50 demonstrations over 1000 epochs. The ACT model takes either three-view or one-view visual observations combined with joint position information as input, and generates action sequences of fixed length $T$. In our experiments, $T$ is set to 100.

\begin{figure}[ht]
    \centering
    \includegraphics[width=\textwidth]{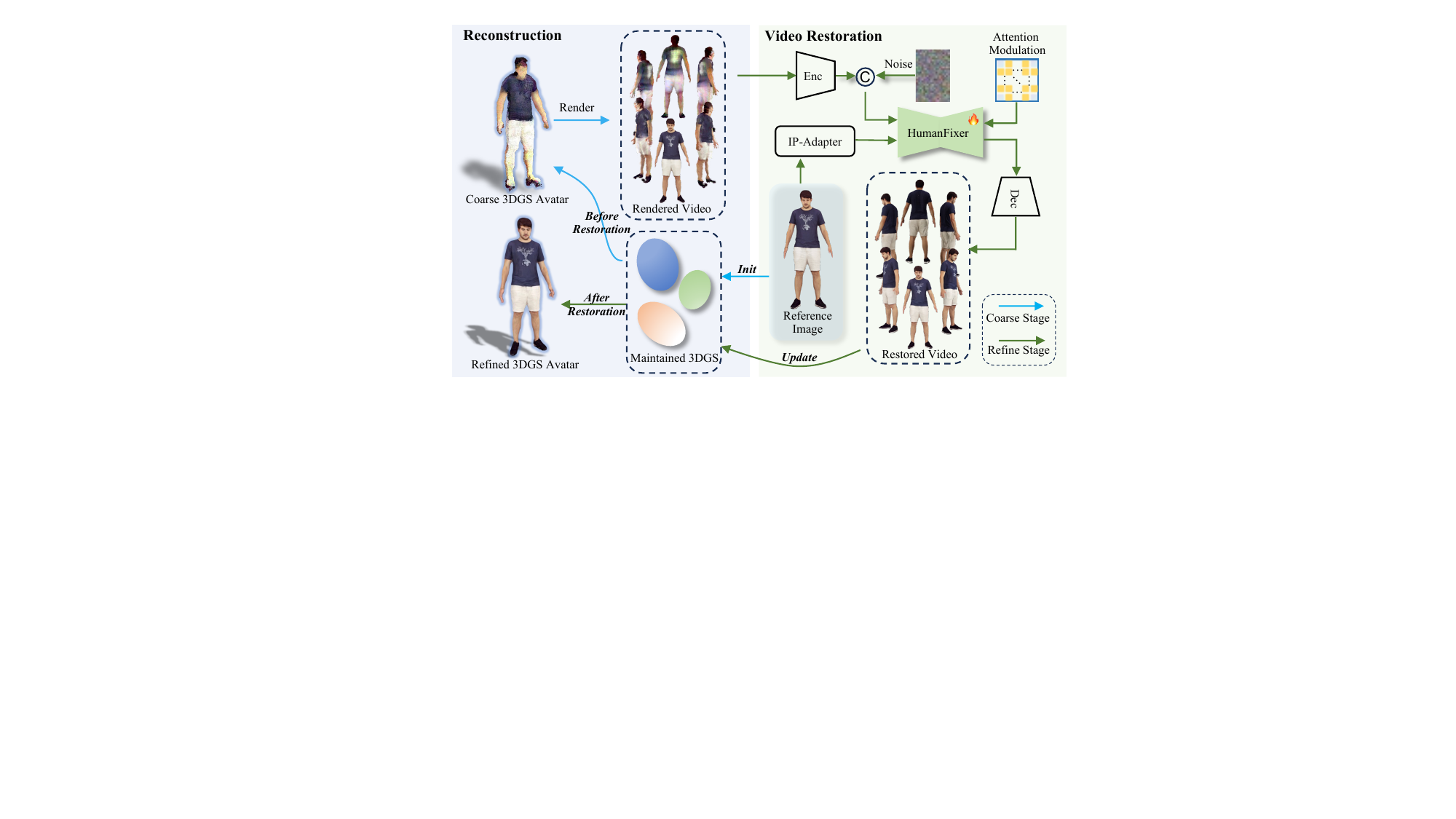}
    \caption{Visualization of two distinct trajectories generated by the policy model from a shared initial frame, after being rendered using \textit{VisAligner}. Despite originating from the same starting image, the trajectories lead to different simulation states due to variations in action sequences. Thanks to the accurate robot positioning in the simulated environment, the resulting photorealistic observations can be effectively used for further policy inference.}
    \label{fig:two_traj}
\end{figure}

For the reinforcement learning phase, the ACT policy receives one-view visual input to produce actions, which are executed in the simulator to generate raw observations. These observations are then processed by \textit{VisAligner} to render photorealistic videos. The last frame of the rendered video serves as the reference image for the next iteration of policy execution. This closed-loop process iterates for a fixed number of steps, yielding trajectories of consistent length with photorealistic visual observations. Two representative trajectories generated by \textit{EmbodieDreamer} are visualized in Figure \ref{fig:two_traj}.

Using this framework, we batch-generate trajectories and apply a preference-based ranking strategy adapted from \cite{grape}. The reward metric is defined as the Euclidean distance between the end-effector and the target position at the terminal state of each trajectory. Based on these reward scores, we select the top-$M$ and bottom-$M$ trajectories as positive and negative samples, respectively. Each fine-tuning cycle trains the ACT model on these $M$ sample pairs for 80 epochs, with the updated policy generating new trajectories for subsequent iterations. We set $M=25$ in all experiments. More details are provided in the Appendix \ref{app:rl}.

We adopt experiments on 4 tasks. Task 1: Grab Paper Cup — success defined as grasping a randomly placed cup. Tasks 2–4 follow a unified structure: (1) \textit{Grasp success} requires picking up the target object (pen, remote, cup), and (2) \textit{Task success} demands precise placement (pen in holder, remote/box, cup/mat), all with randomly positioned objects and targets.

\begin{figure}[ht]
    \centering
    \includegraphics[width=0.8\textwidth]{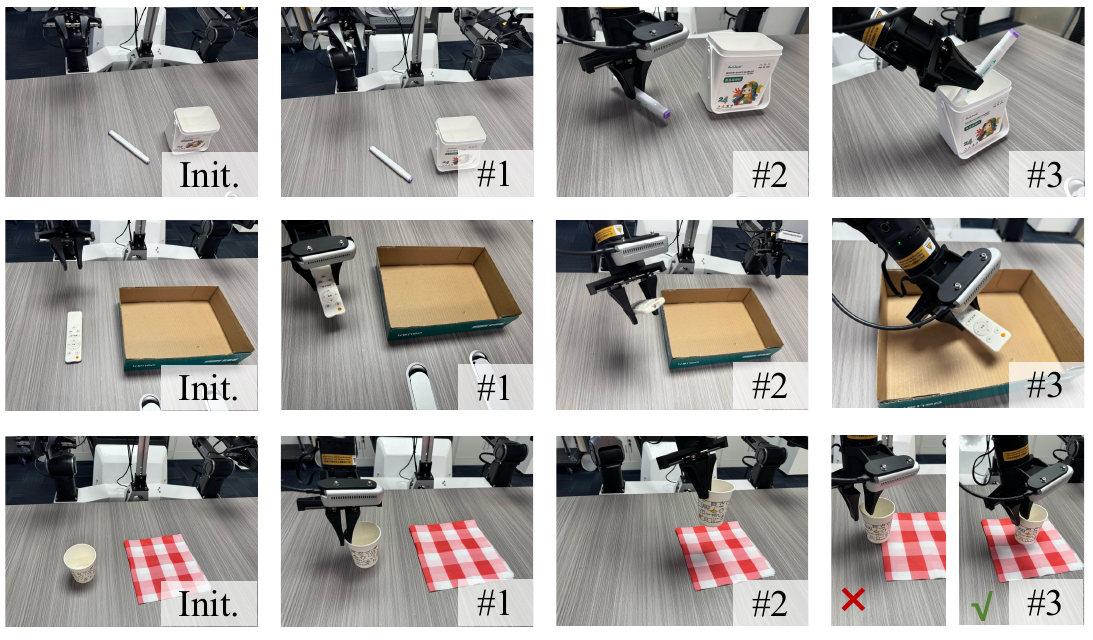}
    \caption{Definition and visualization of the three tasks. Each task is divided into two stages, with success rates reported separately for "grasp success" and "task success". For the \textit{Put Cup on Mat} task, a trial is counted as successful only if the cup is fully placed on the mat.}
    \label{fig:task_demo}
\end{figure}

\paragraph{Experimental Results.} We select four tasks and evaluate their success rates on the real robot system, as shown in Figure~\ref{fig:task_demo}, and the result is shown in Table~\ref{tab:comp_succ_paper}. The experimental results demonstrate that the SFT model performs better with three views compared to a single view. After training with RL, the performance of the single-view model achieves the best results.

\begin{table}[htbp]
\centering
    \caption{We report the success rates of an ACT-based policy across four tasks, under different training settings: (1) trained from scratch with supervised fine-tuning (SFT), (2) without pre-training, and (3) with reinforcement learning (RL) training. The term "with three views" refers to using camera inputs from three angles: the head camera and the wrist cameras on the left and right arms. When reducing the number of views from 3 to 1, a performance drop is observed. However, by incorporating RL training with \textit{EmbodieDreamer} (continuing training from a single view SFT checkpoint), our method achieves superior results compared to other baselines.}
\label{tab:comp_succ_paper}
\resizebox{\linewidth}{!}{
\begin{tabular}{lccccccc|c}
  \toprule
  Method  & Grab Paper Cup & \multicolumn{2}{c}{Put Pen in Cup} & \multicolumn{2}{c}{Put Remote in Box} & \multicolumn{2}{c}{Put Cup on Mat}\\
  \midrule
                       &   Task Suc     & Grasp Suc & Task Suc & Grasp Suc & Task Suc & Grasp Suc & Task Suc & Avg\\
  SFT with three views &  0.7         & 0.5           & 0.5          &  0.7         &      \textbf{0.7}      &    0.6           &      0.5 & 0.60\\
  SFT with one view    &  0.5     & 0.5             &  0.2            &  0.6         &      0.6      &    0.6           &      0.4       & 0.49  \\
  RL with \textit{EmbodieDreamer} & \textbf{0.9} &  \textbf{0.7 }& \textbf{0.7} &\textbf{0.8} & \textbf{0.8} & \textbf{0.8} & \textbf{0.7}  & \textbf{0.77}\\
  \bottomrule
\end{tabular}
}
\end{table}

\subsection{IL with EmbodieDreamer}

\paragraph{Experimental Setup.}
In addition to its applicability in reinforcement learning, \textit{EmbodieDreamer} can also be effectively employed for imitation learning.
We observe that variations in the texture of foreground objects and environmental backgrounds can lead to a noticeable decline in model performance, indicating a certain level of sensitivity to visual appearance changes. To address this issue,We employ \textit{EmbodieDreamer} to generate augmented training data with modified object appearances and environmental conditions, eliminating the need for physical scene configurations or real-world data acquisition. Imitation learning is then applied to train the policy model.

Specifically, we utilize Gemini-2.0-Flash-Preview-Image-Generation \cite{team2023gemini} to modify the foreground and background content of reference images for each episode in the real-robot data (e.g., randomly placing an A4 white paper on the table or altering the color of cups). By replaying the actions from each frame of the original data, we obtain simulated observations under these modified conditions. \textit{EmbodieDreamer} subsequently generates videos under these new conditions, and finally translates these simulated episodes into realistic visual demonstrations, resulting in a set of high-quality, diverse demonstrations by Sim2Real transfer.

To demonstrate the effectiveness of data-augmented imitation learning, we compare three policy configurations:
\begin{itemize}
    \item Model 1: Trained on real data (from scratch, only white cup and no A4 paper)
    \item Model 2: Trained on generated data (from scratch)
    \item Model 3: Trained on generated data (resume from Model1)
\end{itemize}

For background variation, the evaluation includes 50 episodes each from both generated and real data sources. For foreground variation, 50 episodes are taken from real data and 20 from generated data.

\paragraph{Experimental Results.}
We conduct 30 test trials for each task setting, and the success rates are summarized in Table~\ref{tab:augmentation_ablation}. The results show that the use of generated data leads to an improvement in success rate on new scenes, demonstrating the effectiveness of our generated data in facilitating policy transfer. Additionally, we perform post-training on the provided pretrained model $\pi_0$~\cite{pi0} and test its success rate, as shown in Table~\ref{tab:augmentation_ablation_pi0}. The results further validate the utility of the generated data.
\begin{table}[htbp]
\centering
\caption{Success rate comparison of ACT-based policy trained with different data configurations.}
\label{tab:augmentation_ablation}
\resizebox{\linewidth}{!}{
\begin{tabular}{lccc|cccc}
  \toprule
  \multirow{2}{*}{Data Source} & \multirow{2}{*}{Train Method} &Origin& Change Foreground  &Origin& \multicolumn{2}{c}{Change Background} \\
   \cmidrule(lr){3-7}
  & & White Cup & Blue Cup & No A4 & A4 paper(horizontal) & A4 paper(vertical) \\
  \midrule
  Real data & scratch &\textbf{0.467} & 0.333 & \textbf{0.467} & 0.300 & 0.167 \\
  Generated data & scratch & 0.033 & 0.100 & 0.067 &0.567 & 0.500 \\
  Generated data & resume & 0.133 & \textbf{0.433} & 0.133 & \textbf{0.700} & \textbf{0.600} \\
  \bottomrule
\end{tabular}
}
\end{table}

\begin{table}[htbp]
\centering
\caption{Success rate comparison of $\pi_0$ trained with different data configurations.}
\label{tab:augmentation_ablation_pi0}
\resizebox{\linewidth}{!}{
\begin{tabular}{lccc|cccc}
  \toprule
  \multirow{2}{*}{Data Source} &Origin& Change Foreground  &Origin& \multicolumn{2}{c}{Change Background} \\
   \cmidrule(lr){2-6}
  & White Cup & Blue Cup & No A4 & A4 paper(horizontal) & A4 paper(vertical) \\
  \midrule
  Real data & \textbf{0.600} & 0.533  & \textbf{0.600}  & 0.500 & 0.433\\
  Generated data & 0.433 & \textbf{0.633} & 0.433 & \textbf{0.567}& \textbf{0.600}  \\
  \bottomrule
\end{tabular}
}
\end{table}
\section{Conclusion}
This work presents \textit{EmbodieDreamer}, a novel framework addressing the critical Real2Sim2Real gap in Embodied AI by synergistically bridging discrepancies in physical dynamics and visual appearance. Through the proposed \textit{PhysAligner}, we achieve precise alignment of simulated physics with real-world dynamics via differentiable optimization of robot-specific parameters, significantly outperforming conventional methods in both accuracy and efficiency. Concurrently, \textit{VisAligner} leverages conditional video diffusion to synthesize photorealistic environments from low-fidelity simulations, enabling effective visual transfer for policy learning. Experimental results demonstrate that our framework not only reduces physical parameter estimation errors and accelerates optimization but also enhances real-world task success rates through high-fidelity visual rendering. By unifying physics-aware and vision-centric adaptation, \textit{EmbodieDreamer} advances the deployment of simulation-trained policies into real-world scenarios, paving the way for scalable and efficient embodied intelligence development. Future work will explore extending this paradigm to multi-modal sensory integration and long-horizon autonomous tasks.

\textbf{Limitations.} Our current approach has several limitations. First, as a diffusion-based model, the generation of high-quality photorealistic videos is computationally expensive, often taking around two minutes per sequence. While this limitation can be mitigated with faster sampling techniques or network architectures, it still poses challenges for real-time applications. Second, the accuracy of our physics-aware simulation relies heavily on the fidelity of the underlying simulator in modeling real-world dynamics, which may limit generalization to more complex or unstructured environments.

\newpage

\setcounter{section}{0}
In the appendix, we provide more details about \textit{PhysAligner} and present a detailed description of the reinforcement learning procedure with \textit{EmbodieDreamer}. 

\renewcommand{\thetable}{\alph{table}}
\renewcommand{\thefigure}{\alph{figure}}

\section{More Details of PhysAligner}
\label{app:physaligner}
We define the trajectory loss as
\begin{equation}
  \mathcal{L}_{\text{traj}}(f,p,d)
  \;=\;
  \mathcal{L}_{\text{trans}}(f,p,d)
  \;+\;
  \mathcal{L}_{\text{rot}}(f,p,d),
\end{equation}
where $ f $ denotes the input visual observation, $ p $ represents the ground-truth end-effector pose sequence, and $ d $ indicates the diffusion model used for trajectory generation. The loss function jointly optimizes two components: 

\textbf{Translation Loss} $ \mathcal{L}_{\text{trans}} $:  
For each time step $ i \in \{1,2,\dots,T\} $, we compute the Euclidean distance between the predicted position $ \mathbf{x}_i' \in \mathbb{R}^3 $ and the ground-truth position $ \mathbf{x}_i \in \mathbb{R}^3 $ of the end-effector:
\begin{equation}
  \mathcal{L}_{\text{trans}}(f,p,d)
  \;=\;
  \frac{1}{T}
  \sum_{i=1}^{T}
  \lVert \mathbf{x}_i - \mathbf{x}_i' \rVert_2.
\end{equation}

\textbf{Rotation Loss} $ \mathcal{L}_{\text{rot}} $:  
To measure rotational discrepancies, we calculate the Frobenius norm of the difference between the ground-truth rotation matrix $ \mathbf{R}_i \in SO(3) $ and the predicted rotation matrix $ \mathbf{R}_i' \in SO(3) $, normalized by a geometric scaling factor $ 2\sqrt{2} $. The arcsine transformation maps this matrix discrepancy to an angular error in radians:
\begin{equation}
  \mathcal{L}_{\text{rot}}(f,p,d)
  \;=\;
  \frac{1}{T}
  \sum_{i=1}^{T}
  \arcsin\!\Bigl(
    \frac{1}{2\sqrt{2}}\,
    \lVert \mathbf{R}_i - \mathbf{R}_i' \rVert_F
  \Bigr).
\end{equation}

This formulation explicitly models both translational and rotational errors of the end-effector pose $ (\mathbf{x}_i, \mathbf{R}_i) $, ensuring geometric consistency between the simulated trajectory and real-world demonstrations. The $ 2\sqrt{2} $ normalization factor arises from the maximum possible Frobenius norm of $ \mathbf{R}_i - \mathbf{R}_i' $, derived from the properties of orthogonal matrices in $ SO(3) $.

\section{RL Formulation and TPO Loss Adaptation}
\label{app:rl}
To optimize ACT beyond supervised learning, we employ reinforcement learning, following prior work on vision-language alignment~\cite{schulman2017proximal, bai2022training}. We formulate the objective as:

\begin{equation}
\label{eqn:rl_rewrite}
\max_{\pi_\theta} \mathbb{E}_{\zeta \sim \pi_\theta} \left[ r_\phi(\zeta) - \beta \log \frac{\pi_\theta(\zeta)}{\pi_{\text{ref}}(\zeta)} \right],
\end{equation}

where $\pi_{\text{ref}}$ denotes a supervised-trained reference policy, and $\beta$ controls deviation regularization. This form mirrors the KL-regularized RL objective while allowing simplified implementation in trajectory space.

To optimize this objective using trajectory-level preference signals, we follow the principle of Trajectory Preference Optimization (TPO)~\cite{grape}, adapting the loss to our discrete-action scenario. For a pair of trajectories $(\zeta_w, \zeta_l)$ denoting preferred and less-preferred samples, the loss is:

\begin{equation}
\label{eqn:tpo_rewrite}
\mathcal{L}_{\text{TPO}} = - \mathbb{E}_{(\zeta_w, \zeta_l) \sim \mathcal{D}} \left[ \log \sigma\left( \beta \cdot \Delta(\zeta_w, \zeta_l) \right) \right],
\end{equation}

where $\Delta(\zeta_w, \zeta_l)$ represents the log-probability advantage:

\begin{equation}
\label{eqn:delta_def}
\Delta(\zeta_w, \zeta_l) = \log \frac{\pi_\theta(\zeta_w)/\pi_{\text{ref}}(\zeta_w)}{\pi_\theta(\zeta_l)/\pi_{\text{ref}}(\zeta_l)}.
\end{equation}

Assuming a Markovian structure, the log-probability of a trajectory decomposes as:

\begin{equation}
\label{eqn:log_ratio_rewrite}
\log\frac{\pi_\theta(\zeta)}{\pi_{\text{ref}}(\zeta)} = \sum_{t=1}^T \log \frac{\pi_\theta(a_t | o_t)}{\pi_{\text{ref}}(a_t | o_t)}.
\end{equation}

While the standard trajectory likelihood $\pi_\theta(\zeta)$ can be computed exactly in continuous-action models, it becomes impractical in our discrete-action ACT setting. Instead, we propose a tractable formulation tailored to ACT, where the log-likelihood of a trajectory is computed via squared prediction errors:

\begin{equation}
\label{eqn:log_prob_act}
\log \pi(\zeta) = -\frac{1}{2} \sum_{t=1}^{T_\zeta} \| \hat{a}_t - a_t \|^2,
\end{equation}

where $\hat{a}_t$ denotes the predicted action at step $t$ and $a_t$ is the corresponding ground-truth action. This formulation assumes a Gaussian likelihood over discrete action embeddings and reflects the model's trajectory-level confidence.

Using this definition of log-likelihood, the TPO loss becomes:

\begin{equation}
\label{eqn:tpo_custom}
\mathcal{L}_{\text{TPO}} = - \mathbb{E}_{(\zeta_w, \zeta_l) \sim \mathcal{D}} \left[ \log \sigma \left( \beta \cdot \left[ \log \pi_\theta(\zeta_w) - \log \pi_{\text{ref}}(\zeta_w) - \left( \log \pi_\theta(\zeta_l) - \log \pi_{\text{ref}}(\zeta_l) \right) \right] \right) \right].
\end{equation}

This formulation enables stable and efficient preference-based training without requiring explicit probability computation over discrete outputs.
To construct the dataset $\mathcal{D}$, we sample trajectories from a single-view ACT-based policy. These trajectories are ranked using a cost function, and based on their rankings, we categorize them into chosen ($\zeta_w$) and rejected ($\zeta_l$) sets. The TPO loss is computed using these paired trajectories, enabling the policy to learn from preference-aligned data.

\newpage
\normalem
\bibliographystyle{unsrt}
\bibliography{ref}

\begin{thebibliography}{10}

\bibitem{mujuco}
Emanuel Todorov, Tom Erez, and Yuval Tassa.
\newblock Mujoco: A physics engine for model-based control.
\newblock In {\em 2012 IEEE/RSJ International Conference on Intelligent Robots and Systems}, pages 5026--5033. IEEE, 2012.

\bibitem{rlbench}
Stephen James, Zicong Ma, David Rovick~Arrojo, and Andrew~J. Davison.
\newblock Rlbench: The robot learning benchmark \& learning environment.
\newblock {\em IEEE Robotics and Automation Letters}, 2020.

\bibitem{sapien}
Fanbo Xiang, Yuzhe Qin, Kaichun Mo, Yikuan Xia, Hao Zhu, Fangchen Liu, Minghua Liu, Hanxiao Jiang, Yifu Yuan, He~Wang, Li~Yi, Angel~X. Chang, Leonidas~J. Guibas, and Hao Su.
\newblock {SAPIEN}: A simulated part-based interactive environment.
\newblock In {\em The IEEE Conference on Computer Vision and Pattern Recognition (CVPR)}, June 2020.

\bibitem{simplerenv}
Xuanlin Li, Kyle Hsu, Jiayuan Gu, Karl Pertsch, Oier Mees, Homer~Rich Walke, Chuyuan Fu, Ishikaa Lunawat, Isabel Sieh, Sean Kirmani, Sergey Levine, Jiajun Wu, Chelsea Finn, Hao Su, Quan Vuong, and Ted Xiao.
\newblock Evaluating real-world robot manipulation policies in simulation.
\newblock {\em arXiv preprint arXiv:2405.05941}, 2024.

\bibitem{genesis}
Genesis Authors.
\newblock Genesis: A universal and generative physics engine for robotics and beyond, December 2024.

\bibitem{vlabench}
Shiduo Zhang, Zhe Xu, Peiju Liu, Xiaopeng Yu, Yuan Li, Qinghui Gao, Zhaoye Fei, Zhangyue Yin, Zuxuan Wu, Yu-Gang Jiang, and Xipeng Qiu.
\newblock Vlabench: A large-scale benchmark for language-conditioned robotics manipulation with long-horizon reasoning tasks, 2024.

\bibitem{roboverse}
Haoran Geng, Feishi Wang, Songlin Wei, Yuyang Li, Bangjun Wang, Boshi An, Charlie~Tianyue Cheng, Haozhe Lou, Peihao Li, Yen-Jen Wang, Yutong Liang, Dylan Goetting, Chaoyi Xu, Haozhe Chen, Yuxi Qian, Yiran Geng, Jiageng Mao, Weikang Wan, Mingtong Zhang, Jiangran Lyu, Siheng Zhao, Jiazhao Zhang, Jialiang Zhang, Chengyang Zhao, Haoran Lu, Yufei Ding, Ran Gong, Yuran Wang, Yuxuan Kuang, Ruihai Wu, Baoxiong Jia, Carlo Sferrazza, Hao Dong, Siyuan Huang, Yue Wang, Jitendra Malik, and Pieter Abbeel.
\newblock Roboverse: Towards a unified platform, dataset and benchmark for scalable and generalizable robot learning, April 2025.

\bibitem{robotwin}
Yao Mu, Tianxing Chen, Zanxin Chen, Shijia Peng, Zhiqian Lan, Zeyu Gao, Zhixuan Liang, Qiaojun Yu, Yude Zou, Mingkun Xu, et~al.
\newblock Robotwin: Dual-arm robot benchmark with generative digital twins.
\newblock {\em arXiv preprint arXiv:2504.13059}, 2025.

\bibitem{svd}
Andreas Blattmann, Tim Dockhorn, Sumith Kulal, Daniel Mendelevitch, Maciej Kilian, Dominik Lorenz, Yam Levi, Zion English, Vikram Voleti, Adam Letts, et~al.
\newblock Stable video diffusion: Scaling latent video diffusion models to large datasets.
\newblock {\em arXiv preprint arXiv:2311.15127}, 2023.

\bibitem{cogvideo}
Wenyi Hong, Ming Ding, Wendi Zheng, Xinghan Liu, and Jie Tang.
\newblock Cogvideo: Large-scale pretraining for text-to-video generation via transformers.
\newblock {\em arXiv preprint arXiv:2205.15868}, 2022.

\bibitem{humandreamer}
Boyuan Wang, Xiaofeng Wang, Chaojun Ni, Guosheng Zhao, Zhiqin Yang, Zheng Zhu, Muyang Zhang, Yukun Zhou, Xinze Chen, Guan Huang, Lihong Liu, and Xingang Wang.
\newblock Humandreamer: Generating controllable human-motion videos via decoupled generation.
\newblock {\em arXiv preprint arXiv:2503.24026}, 2025.

\bibitem{humandreamerx}
Boyuan Wang, Runqi Ouyang, Xiaofeng Wang, Zheng Zhu, Guosheng Zhao, Chaojun Ni, Guan Huang, Lihong Liu, and Xingang Wang.
\newblock Humandreamer-x: Photorealistic single-image human avatars reconstruction via gaussian restoration.
\newblock {\em arXiv preprint arXiv:2504.03536}, 2025.

\bibitem{rebot}
Yu~Fang, Yue Yang, Xinghao Zhu, Kaiyuan Zheng, Gedas Bertasius, Daniel Szafir, and Mingyu Ding.
\newblock Rebot: Scaling robot learning with real-to-sim-to-real robotic video synthesis, 2025.

\bibitem{roboengine}
Chengbo Yuan, Suraj Joshi, Shaoting Zhu, Hang Su, Hang Zhao, and Yang Gao.
\newblock Roboengine: Plug-and-play robot data augmentation with semantic robot segmentation and background generation.
\newblock {\em arXiv preprint arXiv:2503.18738}, 2025.

\bibitem{irasim}
Fangqi Zhu, Hongtao Wu, Song Guo, Yuxiao Liu, Chilam Cheang, and Tao Kong.
\newblock Irasim: Learning interactive real-robot action simulators.
\newblock {\em arXiv:2406.12802}, 2024.

\bibitem{huang2025dexterous}
Dongchi Huang, Tianle Zhang, Yihang Li, Ling Zhao, Jiayi Li, Zhirui Fang, Chunhe Xia, Lusong Li, and Xiaodong He.
\newblock Dexterous hand manipulation via efficient imitation-bootstrapped online reinforcement learning.
\newblock {\em arXiv preprint arXiv:2503.04014}, 2025.

\bibitem{zare2024survey}
Maryam Zare, Parham~M Kebria, Abbas Khosravi, and Saeid Nahavandi.
\newblock A survey of imitation learning: Algorithms, recent developments, and challenges.
\newblock {\em IEEE Transactions on Cybernetics}, 2024.

\bibitem{tang2025deep}
Chen Tang, Ben Abbatematteo, Jiaheng Hu, Rohan Chandra, Roberto Mart{\'\i}n-Mart{\'\i}n, and Peter Stone.
\newblock Deep reinforcement learning for robotics: A survey of real-world successes.
\newblock In {\em Proceedings of the AAAI Conference on Artificial Intelligence}, volume~39, pages 28694--28698, 2025.

\bibitem{lin2025sim}
Toru Lin, Kartik Sachdev, Linxi Fan, Jitendra Malik, and Yuke Zhu.
\newblock Sim-to-real reinforcement learning for vision-based dexterous manipulation on humanoids.
\newblock {\em arXiv preprint arXiv:2502.20396}, 2025.

\bibitem{pi0}
Kevin Black, Noah Brown, Danny Driess, Adnan Esmail, Michael Equi, Chelsea Finn, Niccolo Fusai, Lachy Groom, Karol Hausman, Brian Ichter, Szymon Jakubczak, Tim Jones, Liyiming Ke, Sergey Levine, Adrian Li-Bell, Mohith Mothukuri, Suraj Nair, Karl Pertsch, Lucy~Xiaoyang Shi, James Tanner, Quan Vuong, Anna Walling, Haohuan Wang, and Ury Zhilinsky.
\newblock $\pi$0: A vision-language-action flow model for general robot control.
\newblock {\em ArXiv}, abs/2410.24164, 2024.

\bibitem{pi05}
Physical Intelligence, Kevin Black, Noah Brown, James Darpinian, Karan Dhabalia, Danny Driess, Adnan Esmail, Michael Equi, Chelsea Finn, Niccolo Fusai, Manuel~Y. Galliker, Dibya Ghosh, Lachy Groom, Karol Hausman, Brian Ichter, Szymon Jakubczak, Tim Jones, Liyiming Ke, Devin LeBlanc, Sergey Levine, Adrian Li-Bell, Mohith Mothukuri, Suraj Nair, Karl Pertsch, Allen~Z. Ren, Lucy~Xiaoyang Shi, Laura Smith, Jost~Tobias Springenberg, Kyle Stachowicz, James Tanner, Quan Vuong, Homer Walke, Anna Walling, Haohuan Wang, Lili Yu, and Ury Zhilinsky.
\newblock $\pi_{0.5}$: a vision-language-action model with open-world generalization, 2025.

\bibitem{team2024octo}
Octo~Model Team, Dibya Ghosh, Homer Walke, Karl Pertsch, Kevin Black, Oier Mees, Sudeep Dasari, Joey Hejna, Tobias Kreiman, Charles Xu, et~al.
\newblock Octo: An open-source generalist robot policy.
\newblock {\em arXiv preprint arXiv:2405.12213}, 2024.

\bibitem{carvalho2023motion}
Joao Carvalho, An~T Le, Mark Baierl, Dorothea Koert, and Jan Peters.
\newblock Motion planning diffusion: Learning and planning of robot motions with diffusion models.
\newblock In {\em 2023 IEEE/RSJ International Conference on Intelligent Robots and Systems (IROS)}, pages 1916--1923. IEEE, 2023.

\bibitem{ze20243d}
Yanjie Ze, Gu~Zhang, Kangning Zhang, Chenyuan Hu, Muhan Wang, and Huazhe Xu.
\newblock 3d diffusion policy: Generalizable visuomotor policy learning via simple 3d representations.
\newblock {\em arXiv preprint arXiv:2403.03954}, 2024.

\bibitem{rafailov2023direct}
Rafael Rafailov, Archit Sharma, Eric Mitchell, Christopher~D Manning, Stefano Ermon, and Chelsea Finn.
\newblock Direct preference optimization: Your language model is secretly a reward model.
\newblock {\em Advances in Neural Information Processing Systems}, 36:53728--53741, 2023.

\bibitem{grape}
Zijian Zhang, Kaiyuan Zheng, Zhaorun Chen, Joel Jang, Yi~Li, Chaoqi Wang, Mingyu Ding, Dieter Fox, and Huaxiu Yao.
\newblock Grape: Generalizing robot policy via preference alignment, 2024.

\bibitem{zhao2023learning}
Tony~Z Zhao, Vikash Kumar, Sergey Levine, and Chelsea Finn.
\newblock Learning fine-grained bimanual manipulation with low-cost hardware.
\newblock {\em arXiv preprint arXiv:2304.13705}, 2023.

\bibitem{brohan2023rt}
Anthony Brohan, Noah Brown, Justice Carbajal, Yevgen Chebotar, Xi~Chen, Krzysztof Choromanski, Tianli Ding, Danny Driess, Avinava Dubey, Chelsea Finn, et~al.
\newblock Rt-2: Vision-language-action models transfer web knowledge to robotic control.
\newblock {\em arXiv preprint arXiv:2307.15818}, 2023.

\bibitem{chi2023diffusion}
Cheng Chi, Zhenjia Xu, Siyuan Feng, Eric Cousineau, Yilun Du, Benjamin Burchfiel, Russ Tedrake, and Shuran Song.
\newblock Diffusion policy: Visuomotor policy learning via action diffusion.
\newblock {\em The International Journal of Robotics Research}, page 02783649241273668, 2023.

\bibitem{rdt1b}
Songming Liu, Lingxuan Wu, Bangguo Li, Hengkai Tan, Huayu Chen, Zhengyi Wang, Ke~Xu, Hang Su, and Jun Zhu.
\newblock Rdt-1b: a diffusion foundation model for bimanual manipulation.
\newblock {\em arXiv preprint arXiv:2410.07864}, 2024.

\bibitem{xue2025reactive}
Han Xue, Jieji Ren, Wendi Chen, Gu~Zhang, Yuan Fang, Guoying Gu, Huazhe Xu, and Cewu Lu.
\newblock Reactive diffusion policy: Slow-fast visual-tactile policy learning for contact-rich manipulation.
\newblock {\em arXiv preprint arXiv:2503.02881}, 2025.

\bibitem{openvla}
{Moo Jin} Kim, Karl Pertsch, Siddharth Karamcheti, Ted Xiao, Ashwin Balakrishna, Suraj Nair, Rafael Rafailov, Ethan Foster, Grace Lam, Pannag Sanketi, Quan Vuong, Thomas Kollar, Benjamin Burchfiel, Russ Tedrake, Dorsa Sadigh, Sergey Levine, Percy Liang, and Chelsea Finn.
\newblock Openvla: An open-source vision-language-action model.
\newblock {\em arXiv preprint arXiv:2406.09246}, 2024.

\bibitem{octo}
{Octo Model Team}, Dibya Ghosh, Homer Walke, Karl Pertsch, Kevin Black, Oier Mees, Sudeep Dasari, Joey Hejna, Charles Xu, Jianlan Luo, Tobias Kreiman, {You Liang} Tan, Pannag Sanketi, Quan Vuong, Ted Xiao, Dorsa Sadigh, Chelsea Finn, and Sergey Levine.
\newblock Octo: An open-source generalist robot policy.
\newblock In {\em Proceedings of Robotics: Science and Systems}, Delft, Netherlands, 2024.

\bibitem{zhong2025dexgraspvla}
Yifan Zhong, Xuchuan Huang, Ruochong Li, Ceyao Zhang, Yitao Liang, Yaodong Yang, and Yuanpei Chen.
\newblock Dexgraspvla: A vision-language-action framework towards general dexterous grasping.
\newblock {\em arXiv preprint arXiv:2502.20900}, 2025.

\bibitem{zhao2025cot}
Qingqing Zhao, Yao Lu, Moo~Jin Kim, Zipeng Fu, Zhuoyang Zhang, Yecheng Wu, Zhaoshuo Li, Qianli Ma, Song Han, Chelsea Finn, et~al.
\newblock Cot-vla: Visual chain-of-thought reasoning for vision-language-action models.
\newblock {\em arXiv preprint arXiv:2503.22020}, 2025.

\bibitem{bridgedatav2}
Homer Walke, Kevin Black, Abraham Lee, Moo~Jin Kim, Max Du, Chongyi Zheng, Tony Zhao, Philippe Hansen-Estruch, Quan Vuong, Andre He, Vivek Myers, Kuan Fang, Chelsea Finn, and Sergey Levine.
\newblock Bridgedata v2: A dataset for robot learning at scale.
\newblock In {\em Conference on Robot Learning (CoRL)}, 2023.

\bibitem{rt1}
Anthony Brohan, Noah Brown, Justice Carbajal, Yevgen Chebotar, Joseph Dabis, Chelsea Finn, Keerthana Gopalakrishnan, Karol Hausman, Alex Herzog, Jasmine Hsu, Julian Ibarz, Brian Ichter, Alex Irpan, Tomas Jackson, Sally Jesmonth, Nikhil Joshi, Ryan Julian, Dmitry Kalashnikov, Yuheng Kuang, Isabel Leal, Kuang-Huei Lee, Sergey Levine, Yao Lu, Utsav Malla, Deeksha Manjunath, Igor Mordatch, Ofir Nachum, Carolina Parada, Jodilyn Peralta, Emily Perez, Karl Pertsch, Jornell Quiambao, Kanishka Rao, Michael Ryoo, Grecia Salazar, Pannag Sanketi, Kevin Sayed, Jaspiar Singh, Sumedh Sontakke, Austin Stone, Clayton Tan, Huong Tran, Vincent Vanhoucke, Steve Vega, Quan Vuong, Fei Xia, Ted Xiao, Peng Xu, Sichun Xu, Tianhe Yu, and Brianna Zitkovich.
\newblock Rt-1: Robotics transformer for real-world control at scale.
\newblock In {\em arXiv preprint arXiv:2212.06817}, 2022.

\bibitem{magicdrive}
Ruiyuan Gao, Kai Chen, Enze Xie, Lanqing Hong, Zhenguo Li, Dit-Yan Yeung, and Qiang Xu.
\newblock {MagicDrive}: Street view generation with diverse 3d geometry control.
\newblock In {\em International Conference on Learning Representations}, 2024.

\bibitem{drivedreamer}
Xiaofeng Wang, Zheng Zhu, Guan Huang, Xinze Chen, Jiagang Zhu, and Jiwen Lu.
\newblock Drivedreamer: Towards real-world-driven world models for autonomous driving.
\newblock {\em arXiv preprint arXiv:2309.09777}, 2023.

\bibitem{worlddreamer}
Xiaofeng Wang, Zheng Zhu, Guan Huang, Boyuan Wang, Xinze Chen, and Jiwen Lu.
\newblock Worlddreamer: Towards general world models for video generation via predicting masked tokens.
\newblock {\em arXiv preprint arXiv:2401.09985}, 2024.

\bibitem{gaia}
Anthony Hu, Lloyd Russell, Hudson Yeo, Zak Murez, George Fedoseev, Alex Kendall, Jamie Shotton, and Gianluca Corrado.
\newblock Gaia-1: A generative world model for autonomous driving.
\newblock {\em arXiv preprint arXiv:2309.17080}, 2023.

\bibitem{drivedreamer4d}
Guosheng Zhao, Chaojun Ni, Xiaofeng Wang, Zheng Zhu, Guan Huang, Xinze Chen, Boyuan Wang, Youyi Zhang, Wenjun Mei, and Xingang Wang.
\newblock Drivedreamer4d: World models are effective data machines for 4d driving scene representation.
\newblock {\em arXiv preprint arXiv:2410.13571}, 2024.

\bibitem{streetgaussian}
Yunzhi Yan, Haotong Lin, Chenxu Zhou, Weijie Wang, Haiyang Sun, Kun Zhan, Xianpeng Lang, Xiaowei Zhou, and Sida Peng.
\newblock Street gaussians for modeling dynamic urban scenes.
\newblock {\em arXiv preprint arXiv:2401.01339}, 2024.

\bibitem{recondreamer}
Chaojun Ni, Guosheng Zhao, Xiaofeng Wang, Zheng Zhu, Wenkang Qin, Guan Huang, Chen Liu, Yuyin Chen, Yida Wang, Xueyang Zhang, et~al.
\newblock Recondreamer: Crafting world models for driving scene reconstruction via online restoration.
\newblock {\em arXiv preprint arXiv:2411.19548}, 2024.

\bibitem{acdc}
Po-Chen Ko, Jiayuan Mao, Yilun Du, Shao-Hua Sun, and Joshua~B. Tenenbaum.
\newblock Learning to act from actionless videos through dense correspondences, 2023.

\bibitem{unipi}
Yilun Du, Sherry Yang, Bo~Dai, Hanjun Dai, Ofir Nachum, Josh Tenenbaum, Dale Schuurmans, and Pieter Abbeel.
\newblock Learning universal policies via text-guided video generation.
\newblock {\em Advances in neural information processing systems}, 36:9156--9172, 2023.

\bibitem{robodreamer}
Siyuan Zhou, Yilun Du, Jiaben Chen, Yandong Li, Dit-Yan Yeung, and Chuang Gan.
\newblock {R}obo{D}reamer: Learning compositional world models for robot imagination.
\newblock In Ruslan Salakhutdinov, Zico Kolter, Katherine Heller, Adrian Weller, Nuria Oliver, Jonathan Scarlett, and Felix Berkenkamp, editors, {\em Proceedings of the 41st International Conference on Machine Learning}, volume 235 of {\em Proceedings of Machine Learning Research}, pages 61885--61896. PMLR, 21--27 Jul 2024.

\bibitem{manipdreamer}
Ying Li, Xiaobao Wei, Xiaowei Chi, Yuming Li, Zhongyu Zhao, Hao Wang, Ningning Ma, Ming Lu, and Shanghang Zhang.
\newblock Manipdreamer: Boosting robotic manipulation world model with action tree and visual guidance, 2025.

\bibitem{robotransfer}
Liu Liu, Xiaofeng Wang, Guosheng Zhao, Keyu Li, Wenkang Qin, Jiaxiong Qiu, Zheng Zhu, Guan Huang, and Zhizhong Su.
\newblock Robotransfer: Geometry-consistent video diffusion for robotic visual policy transfer, 2025.

\bibitem{3dgs}
Bernhard Kerbl, Georgios Kopanas, Thomas Leimk{\"u}hler, and George Drettakis.
\newblock 3d gaussian splatting for real-time radiance field rendering.
\newblock {\em ACM ToG}, 2023.

\bibitem{nerf}
Ben Mildenhall, Pratul~P Srinivasan, Matthew Tancik, Jonathan~T Barron, Ravi Ramamoorthi, and Ren Ng.
\newblock Nerf: Representing scenes as neural radiance fields for view synthesis.
\newblock {\em Communications of the ACM}, 2021.

\bibitem{recondreamerplus}
Guosheng Zhao, Xiaofeng Wang, Chaojun Ni, Zheng Zhu, Wenkang Qin, Guan Huang, and Xingang Wang.
\newblock Recondreamer++: Harmonizing generative and reconstructive models for driving scene representation.
\newblock {\em arXiv preprint arXiv:2503.18438}, 2025.

\bibitem{wonderturbo}
Chaojun Ni, Xiaofeng Wang, Zheng Zhu, Weijie Wang, Haoyun Li, Guosheng Zhao, Jie Li, Wenkang Qin, Guan Huang, and Wenjun Mei.
\newblock Wonderturbo: Generating interactive 3d world in 0.72 seconds.
\newblock {\em arXiv preprint arXiv:2504.02261}, 2025.

\bibitem{zhou2025gs}
Zelin Zhou, Saurav Uprety, Shichuang Nie, and Hongzhou Yang.
\newblock Gs-gvins: A tightly-integrated gnss-visual-inertial navigation system augmented by 3d gaussian splatting.
\newblock {\em arXiv preprint arXiv:2502.10975}, 2025.

\bibitem{10902412}
Yifan Liu, Chenxin Li, Hengyu Liu, Chen Yang, and Yixuan Yuan.
\newblock Foundation model-guided gaussian splatting for 4d reconstruction of deformable tissues.
\newblock {\em IEEE Transactions on Medical Imaging}, 44(6):2672--2682, 2025.

\bibitem{wang2024og}
Meng Wang, Junyi Wang, Changqun Xia, Chen Wang, and Yue Qi.
\newblock Og-mapping: Octree-based structured 3d gaussians for online dense mapping.
\newblock {\em arXiv preprint arXiv:2408.17223}, 2024.

\bibitem{robo-gs}
Haozhe Lou, Yurong Liu, Yike Pan, Yiran Geng, Jianteng Chen, Wenlong Ma, Chenglong Li, Lin Wang, Hengzhen Feng, Lu~Shi, Liyi Luo, and Yongliang Shi.
\newblock Robo-gs: A physics consistent spatial-temporal model for robotic arm with hybrid representation, 2024.

\bibitem{splatsim}
Mohammad~Nomaan Qureshi, Sparsh Garg, Francisco Yandun, David Held, George Kantor, and Abhishesh Silwal.
\newblock Splatsim: Zero-shot sim2real transfer of rgb manipulation policies using gaussian splatting, 2024.

\bibitem{vae}
Diederik~P Kingma and Max Welling.
\newblock Auto-encoding variational bayes, 2022.

\bibitem{grounding}
Shilong Liu, Zhaoyang Zeng, Tianhe Ren, Feng Li, Hao Zhang, Jie Yang, Chunyuan Li, Jianwei Yang, Hang Su, Jun Zhu, et~al.
\newblock Grounding dino: Marrying dino with grounded pre-training for open-set object detection.
\newblock {\em arXiv preprint arXiv:2303.05499}, 2023.

\bibitem{sam2}
Nikhila Ravi, Valentin Gabeur, Yuan-Ting Hu, Ronghang Hu, Chaitanya Ryali, Tengyu Ma, Haitham Khedr, Roman Rädle, Chloe Rolland, Laura Gustafson, Eric Mintun, Junting Pan, Kalyan~Vasudev Alwala, Nicolas Carion, Chao-Yuan Wu, Ross Girshick, Piotr Dollár, and Christoph Feichtenhofer.
\newblock Sam 2: Segment anything in images and videos, 2024.

\bibitem{mobile_aloha}
Zipeng Fu, Tony~Z. Zhao, and Chelsea Finn.
\newblock Mobile aloha: Learning bimanual mobile manipulation with low-cost whole-body teleoperation.
\newblock In {\em {Conference on Robot Learning (CoRL)}}, 2024.

\bibitem{fvd}
Thomas Unterthiner, Sjoerd Van~Steenkiste, Karol Kurach, Raphael Marinier, Marcin Michalski, and Sylvain Gelly.
\newblock Towards accurate generative models of video: A new metric \& challenges.
\newblock {\em arXiv preprint arXiv:1812.01717}, 2018.

\bibitem{lpips}
Richard Zhang, Phillip Isola, Alexei~A Efros, Eli Shechtman, and Oliver Wang.
\newblock The unreasonable effectiveness of deep features as a perceptual metric.
\newblock In {\em Proceedings of the IEEE conference on computer vision and pattern recognition}, pages 586--595, 2018.

\bibitem{act}
Tony~Z Zhao, Vikash Kumar, Sergey Levine, and Chelsea Finn.
\newblock Learning fine-grained bimanual manipulation with low-cost hardware.
\newblock {\em arXiv preprint arXiv:2304.13705}, 2023.

\bibitem{team2023gemini}
Gemini Team, Rohan Anil, Sebastian Borgeaud, Jean-Baptiste Alayrac, Jiahui Yu, Radu Soricut, Johan Schalkwyk, Andrew~M Dai, Anja Hauth, Katie Millican, et~al.
\newblock Gemini: a family of highly capable multimodal models.
\newblock {\em arXiv preprint arXiv:2312.11805}, 2023.

\bibitem{schulman2017proximal}
John Schulman, Filip Wolski, Prafulla Dhariwal, Alec Radford, and Oleg Klimov.
\newblock Proximal policy optimization algorithms.
\newblock {\em arXiv preprint arXiv:1707.06347}, 2017.

\bibitem{bai2022training}
Yuntao Bai, Andy Jones, Kamal Ndousse, Amanda Askell, Anna Chen, Nova DasSarma, Dawn Drain, Stanislav Fort, Deep Ganguli, Tom Henighan, et~al.
\newblock Training a helpful and harmless assistant with reinforcement learning from human feedback.
\newblock {\em arXiv preprint arXiv:2204.05862}, 2022.

\end{thebibliography}

\end{document}